\documentclass[10pt]{article}
\usepackage[utf8]{inputenc}
\usepackage[letterpaper]{geometry}
\usepackage{hicss}
\usepackage{times}
\usepackage[none]{hyphenat}
\usepackage{url}
\usepackage{latexsym}
\usepackage{indentfirst}
\usepackage{graphicx}
\graphicspath{{images/}}
\usepackage[
    style=apa,
  ]{biblatex}
\usepackage[linewidth=1pt]{mdframed}
\usepackage{graphicx}
\usepackage{amsmath}
\usepackage{amssymb}
\usepackage{amsfonts}
\usepackage{bbm}
\usepackage{xcolor}
\usepackage{algorithm}
\usepackage{algorithmic}
\usepackage{bbding}
\usepackage{pifont}
\usepackage{multirow}
\usepackage{booktabs}
\usepackage{amsthm,multicol,enumerate}
\usepackage{makecell}
\usepackage{color,xcolor}
\usepackage{subcaption}
\usepackage{enumitem}
\usepackage{soul}
\usepackage{array}
\usepackage{multirow}
\usepackage{adjustbox}
\usepackage{tabularx}
\usepackage{etoolbox}
\usepackage{pgf}
\usepackage{etoc}
\usepackage[T1]{fontenc}

\usepackage{microtype}

\usepackage{inconsolata}

\title{Improving Stability Estimates in Adversarial Explainable AI through Alternate Search Methods}

\author{Christopher Burger \\
 University of Mississippi \\
 {\underline{cburger@olemiss.edu}} \And
 Charles Walter \\
 University of Mississippi \\
 {\underline{cwwalter@olemiss.edu}}
 }

\date{}

\begin{document}
\maketitle
\begin{abstract}
Advances in the effectiveness of machine learning models have come at the cost of enormous complexity resulting in a poor understanding of how they function. Local surrogate methods have been used to approximate the workings of these complex models, but recent work has revealed their vulnerability to adversarial attacks where the explanation produced is appreciably different while the meaning and structure of the complex model’s output remains similar. This prior work has focused on the existence of these weaknesses but not on their magnitude. Here we explore using an alternate search method with the goal of finding minimum viable perturbations, the fewest perturbations necessary to achieve a fixed similarity value between the original and altered text’s explanation. Intuitively, a method that requires fewer perturbations to expose a given level of instability is inferior to one which requires more. This nuance allows for superior comparisons of the stability of explainability methods.

\end{abstract}

\subsubsection*{Keywords:}

Explainability, Interpretability, XAI, Stability, Robustness

\section{Introduction}

Recent progress in machine learning and AI has enabled the widespread use of powerful but highly complex models in many aspects of our lives. These advanced models are applied across various fields but in areas with high consequences for failure, such as medicine or finance, there has been slower adoption due to the potential cost of error. All models come with some degree of error, but assigning blame for failures can be challenging, depending on the type of failure. Error within a generative model creating an email template may only cause confusion or embarrassment but failure in a model designed for producing a medical diagnosis is far more serious. 

When any model fails, its stakeholders need to understand the reasons behind the cause of the failure. However, as models grow increasingly complex, directly interpreting their actions becomes impractical. In response, the field of explanatory AI (XAI) aims to develop tools that assist both developers and users to comprehend the workings of a model. One common XAI approach is to create an interpretable approximation of the complex model, known as a surrogate model. These surrogate models help explain why a model made certain predictions and, in practice, focus primarily on explaining individual predictions rather than attempting to explain the entire model. As surrogate models are models themselves, they are subject to the same weaknesses as the original models they are designed to explain.

One necessary property these surrogate models need to satisfy is stability, also known as robustness. Intuitively, a stable model possesses the attribute that when an insignificant change is applied to the input, the corresponding output is subject to at most an insignificant change \parencite{alvarezmelis2018robustness}. A lack of stability is leveraged in adversarial attacks where, given an unstable model, we see changes that should not be meaningful to an ideal model produce appreciably erroneous results. Many common local surrogate models have been shown to lack stability across the major data modalities: image, tabular, and text \parencite{slack2020fooling,ivankay2022fooling}. For XAI specifically, this instability is with respect the explanation provided by the surrogate model. Generally, these explanations are a list of features ordered by importance and, when subject to a successful adversarial attack, the explanation produced is different while the input to the model being explained remains similar. Evidence of instability means the explanation, and the explanatory model itself, cannot be trusted. The original complex model then remains a black-box and may be prevented from use due to lack of understanding surrounding its function due to legal or social conventions despite gains in effectiveness over other models.


Prior work in adversarial XAI has focused on establishing the \textit{existence} of instability and has yet to explore how much instability may be present. Previous work determined the presence of instability using a fixed maximum amount of change applied to the input. Notably, this contrasts to standard adversarial attacks on classification models. Standard attacks often also possess some perturbation or query limit, but the process itself has an inherent end goal which is the change in predicted class. Adversarial attacks on XAI methods lack this innate measure of success and so have relied on search exhaustion or perturbation maximums instead \parencite{xaifooler}. 

As the existence of instability has already been demonstrated, we instead seek to refine our knowledge of it by asking: What is the minimum number of perturbations that will sufficiently alter an explanation. To do so, we need to establish some metric of success, just as that which exists in classical adversarial perturbations. However, unlike a standard adversarial attack in which the alteration of the predicted class is the obvious metric of success, exactly what determines a suitable difference between explanations is more subjective. With a defined threshold, we will be able to answer the question of stability beyond its existence and instead answer how substantial is it, especially compared to other methods. A method that consistently lacks stability when exposed to only a small number of perturbations is clearly worse than one that that requires a consistently large subset of the text to be changed. In Table \ref{tab:minimal_perturbations} we see the first document (26.5\% similarity) requires substantially more perturbations to achieve a comparable level of similarity to that of the second attack (28.9\% similarity). The presence of a single perturbation that provides a significant decrease in similarity implies a much more serious instability when compared to a method that requires multiple perturbations.

We restrict our focus to XAI models for text-based data. Our reasons are as follows: (1) Text-based models are ubiquitous as natural language is a primary communication mechanism. (2) As natural language is woven into the fabric of our daily experiences we as humans have an inherently robust sense for understanding the meaning of a document and so can judge the quality of a text document's explanation effectively without any formal procedure or training. This innate ability to judge the quality of an explanation allows us to choose reasonable thresholds, reducing some of what is an inherently subjective decision. (3) Our depth of investigation can be increased significantly. Testing XAI stability is a computationally intensive process, so focusing on one data type allows a much broader range of thresholds and similarity measures to be tested given a fixed amount of time and computational resources.

To locate these sets of minimal perturbations we test an alternative search process, a genetic algorithm, to allow a wider exploration of the search space. Existing search processes are highly greedy due a combination of being developed primarily to show the existence of instability and due to extensive computational requirements. These processes, while effective, are not optimal for the purposes of investigating just how little perturbation is necessary to achieve a given threshold of instability. The use of the genetic algorithm will then allow more accurate estimates of XAI stability.

\begin{table}[htb]
\caption{Perturbed explanations with close similarity values (calculated with respect to similarity measure RBO$_{0.5}$) despite substantial differences in perturbation rate.}
    \footnotesize
    \begin{subtable}[t]{.48\textwidth}
        \caption*{}
        \raggedright
        \setlength\tabcolsep{4.75 pt}
            \begin{tabular}{clc|lc|lc}
                \hline \\
               & \multicolumn{2}{c}{\textbf{Original}} & \multicolumn{2}{c}{\textbf{26.5\% Sim.}}&  \multicolumn{2}{c}{\textbf{28.9\% Sim.}} \vspace{2pt}
                \\
                 & \multicolumn{1}{c}{\textbf{Feature}} & \multicolumn{1}{c}{\textbf{Weight}}&  \multicolumn{1}{c}{\textbf{Feature}} & \multicolumn{1}{c}{\textbf{Weight}} & \multicolumn{1}{c}{\textbf{Feature}} & \multicolumn{1}{c}{\textbf{Weight}} \\
                \hline \\
                1 & dogs & 2.60 & tennis & 3.16 & tennis & 3.04 \\
                2 & balls & 2.10 & dogs & 2.65 & dogs & 2.68 \\
                3 & helpful & 0.44 & balls & 1.61 & balls & 2.22 \\
                4 & fetching & 0.39 & adore & 0.69 & love & 0.47 \\
                5 & love & 0.29 & useful & 0.49 & fetching & 0.42 \\
                6 & play & 0.07 & toy & 0.34 & toy & 0.39 \\
                7 & tennis & 0.04 & wish & 0.03 & helpful & 0.39 \\
                8 & wish & 0.03 & fetches & 0.02 & wish & 0.09 \\
                \hline \\
            \end{tabular}
    \end{subtable}%
    
     \begin{center}
     \textbf{Original Text}   
    \end{center}
    
    i love dogs ! though i wish mine was more helpful while i play tennis . fetching balls . . .
    \newline

        \begin{center}
     \textbf{Perturbed Text - 26.5\% Similarity}
    \end{center}

    i \textbf{\textit{adore}} dogs ! though i wish mine was more \textbf{\textit{useful}} while i \textbf{\textit{toy}} tennis . \textbf{\textit{fetches}} balls . . .
    \newline

        \begin{center}
     \textbf{Perturbed Text - 28.9\% Similarity}
    \end{center}

    i love dogs ! though i wish mine was more helpful while i \textbf{\textit{toy}} tennis . fetching balls . . .
    
\vspace{5pt}
\hrulefill
\label{tab:minimal_perturbations}
\end{table}

\section{Background \& Related Work}

Prior work on XAI stability has emphasized on evaluating models using tabular or image data across various interpretation methods, which often use small perturbations to the input data to generate appreciably different explanations~\parencite{alvarezmelis2018robustness,InterpretationNN,alvarezmelis2018robustness}, or generate explanations that consist of arbitrary features~\parencite{slack2020fooling}. Our focus here is restricted to the least explored domain, text. Prior work exists directly involving adversarial perturbations for XAI but has been focused on determining the existence of such perturbations rather than establishing  which components in the XAI method are most vulnerable \parencite{sinha2021perturbing,ivankay2022fooling,xaifooler}.

While there are many XAI methods available, we narrow our choices by selecting for three important criteria: proven usage in critical applications, a level of generalizability to other XAI methods, and the explanations generated are to satisfy certain attributes that constitute an effective explanation. Namely concision, order, and weight \parencite{xaifooler}. From these criteria we choose Local Interpretable Model-agnostic Explanations (\textsc{Lime})~\parencite{LIME_Ribeiro} as our target explanatory algorithm 
\textsc{Lime} is a commonly used and referenced tool in XAI frameworks, which has been integrated into critical ML applications such as finance~\parencite{gramegna2021shap} and healthcare~\parencite{kumarakulasinghe2020evaluating,fuhrman2022review}. To explain a prediction, \textsc{Lime} trains a shallow, inherently explainable surrogate model such as Logistic Regression on training examples that are synthesized within a neighborhood of an individual prediction. The resulting explanation is an ordered collection of features and their weights from this surrogate model that satisfies our requirements for a quality explanation outlined above. For text data, explanations generated by \textsc{Lime} have features that are individual words contained within the document to be explained, which can be easily understood even by non-specialists. 

A simplified overview of \textsc{Lime} is given below:

Let $d$ be a document whose prediction under a target model $f(\cdot)$ is to be explained. \textsc{Lime}'s explanation generation for $f(d)$ is as follows. 
\begin{enumerate}
    \item Generate perturbed document $d_i$ by randomly selecting $k$ words from $d$ and removing all of their occurrences.
    \item Repeat \textit{Step 1} to sample $n$ different perturbations $\{d_i\}_{i=1}^{n}$.
    \item Train an explainable model $g(\cdot)$ via supervised learning with features $d_i$ and labels $f(d_i)$.
\end{enumerate}

From the generated explainable model \textsc{Lime} derives the explanation, which is a list of features ranked by their weights. This list ensures out criteria of generalizability, as our method is applicable to any XAI method that returns an explanation in this format. 

\textsc{Lime} has been extensively analyzed for its efficacy. Garreau et al. first investigated the stability of \textsc{Lime} for tabular data \parencite{garreau2020explaining,garreau2022looking}, which showed that important features can be omitted from the resulting explanations by changing parameters and that artifacts of the explanation generation process could ultimately produce misleading explanations. They further extended the analysis to text data \parencite{mardaoui2021analysis} but only with respect to fidelity instead of stability of surrogate models. Other relevant works in text domain include~\cite{ivankay2022fooling}, which utilized gradient-based approach but assumed white-box access to the target model; and \cite{sinha2021perturbing}, which revealed that \textsc{Lime}'s explanations are unstable to black-box text perturbations. However, their experiment design may have led to an underestimation of \textsc{Lime}'s stability as explored in \cite{xaifooler} which investigated the inherent instability of \textsc{Lime}'s sampling process for text data as well as provided an alternate search strategy focused on the preferential perturbation of features deemed unimportant.
 
The previous investigations into \textsc{Lime}'s instability have
followed a procedure similar to that of popular adversarial perturbation algorithm for text classifiers, TextFooler \parencite{Textfooler}. In TextFooler, words are replaced with their nearest neighbors using a chosen embedding space subject to constraints on the part of speech and overall semantic similarity to ensure that the meaning of the original text is maintained as closely as possible. Unlike the default TextFooler attack, whose goal is to alter the classification result and so has a natural endpoint, the attack process on \textsc{Lime} does not have such a clearly defined condition, as the goal is to alter the explanations "enough" to be different while retaining the meaning and the original prediction of the model that is being explained. This "enough" is subjective and so was avoided in prior work as the goal was the demonstration of the existence of instability. Instead, limits based on a maximum number of perturbations, usually, a proportion of the original document length, were employed to provide an unambiguous termination point.

 We note that \textsc{Lime} here is not so much the focus of our inquiry but used as a familiar standard to due to its substantial base of prior work and common use. Since prior work considers only if the perturbed input's explanation is sufficiently different at the \textit{end} of a process, often consisting of many perturbations, we now ask, what would be reasonable similarity thresholds for stopping this process early, and given these thresholds what are the minimal amounts of perturbation needed to reach them?    

\section{Problem Formulation}

As our goal is fundamentally similar to prior work in that we seek the discovery of perturbations that induce instability our constraints when generating adversarial perturbations are generally equivalent to the prior work that has established the existence of \textsc{Lime}'s instability. Namely, the \textit{meaning} of the input is retained, as well as an identical predicted class for the perturbed input under the original model. What follows is summary of the process for generating perturbations and the general search process.

\subsection{Perturbation Strategy}\label{sec:perturbation_strategy}
Let $d_b$ be the original input document and $d_p$ be its perturbed version. For their respective explanations $e_{d_p}$ and $e_{d_b}$ we minimize their explanation similarity or alternatively maximize the distance between them if using a distance function. For simplicity we will refer to any such function in terms of similarity and work to minimize it between the two explanations:
\begin{equation}\label{max_exp_sim}
    d_p = \underset{d_p}{\arg\!\min} \;\; \mathbf{\textbf{\textit{Sim}}_e}(e_{d_b},e_{d_p}),
\end{equation}
\noindent where $\mathbf{\textbf{\textit{Sim}}_e}(e_{d_b},e_{d_p})$ is the similarity function between the two explanations. To optimize Equation (\ref{max_exp_sim}), the methods of previous works involve a series of successive perturbations within the original document as commonly proposed in adversarial text literature. In a typical adversarial text setting, malicious actors aim to manipulate the target model's prediction on the perturbed document, which naturally leads to significant changes in the original explanation. But this is not meaningful for attacks on explanations as substantial changes in the perturbed document tend to lead to changes of similar magnitude in the explanation; thus, we wish to preserve the original prediction while altering only its explanation:
\begin{equation}\label{prediction_unchanged}
    f(d_b) = f(d_p),
\end{equation}
Where $f(\cdot)$ is the model being explained. We note that an alternative question can be proposed, where we seek to induce substantial changes in a document with the goal of minimizing a change in explanation. We continue to use the prior approach to XAI stability in prior work as to best provide a means of comparison against existing literature for our results.

Now changing arbitrarily chosen words with equally arbitrary substitutions will eventually produce an explanation different from the original. However, this will likely produce a perturbed document $d_p$ whose meaning significantly differs from $d_b$. Thus, we impose a constraint on the semantic similarity between $d_b$ and $d_p$ to ensure that the perturbed document $d_p$ does not alter the fundamental meaning of $d_b$:
\begin{equation}\label{semantic_constraint}
\mathbf{\textbf{\textit{Sim}}_s}(d_b,d_p) \geq \delta,
\end{equation}

\noindent where $\mathbf{\textbf{\textit{Sim}}_s}(\cdot)$ is the semantic similarity between two documents and $\delta$ is a sufficiently large hyper-parameter threshold. This threshold is generally not invoked during the search process due to already extensive computational requirements, but serves as check after completion to ensure that the perturbed document is reasonably close to the original input.

Even with the semantic constraint in Equation (\ref{semantic_constraint}), there is usually some degradation in the context and semantics between $d_b$ and $d_p$. Ideally, we want the perturbed document $d_p$ to closely resemble its original $d_b$. However, as the number of perturbations grows larger, the document retains less of its original context and meaning. Prior work addressed this issue by imposing a maximum number of perturbations, $\epsilon$, chosen based on the total length of the document.

\begin{equation} \label{constraint_3}
i \leq \epsilon * |d_b|,
\end{equation}
\noindent where an accepted $i$-th perturbation will have replaced $i$ total number of words (as each perturbation replaces a single word) and $|d_b|$ is the total number of words in $d_b$. 

Here our method diverges from the prior work, while we retain a maximum amount of perturbations, we also use a threshold of similarity, $\tau$ from the measure used to compare the explanations. Where as long as available perturbations subject to the maximum limit exist and the best candidate similarity is above the threshold $\tau$ the search process continues.

\begin{equation} \label{constraint_5}
\mathbf{\textbf{\textit{Sim}}_e}(e_{d_b},e_{d_p}) > \tau ,
\end{equation}

Additionally, we enact the restriction on perturbing top-k elements from the original explanation $e_{d_b}$ from Burger et al. (\citeyear{xaifooler}). Changing a single highly ranked feature can drastically alter the resulting similarity between explanations, even if the word replacement is a direct synonym. Similarity measures with a weighting component are vulnerable to this, and while measures exist that assign what is effectively equal weight to all features, these discard important information within the explanation and should be considered sub-optimal in comparison to weighted measures. 


And so, the set of top $k$ feature(s) belonging to the base explanation $e_{d_b}$ must appear in the perturbed explanation $e_{d_p}$:
\begin{equation} \label{constraint_4}
e_{d_p} \cap c \: \neq \: \varnothing\;\;\; \forall \: c \in e_{d_b}[:k].
\end{equation}

\noindent Then, our objective function is as follows.
\begin{center}
\fbox{\parbox[t]{0.95\linewidth}{
\textbf{\textsc{Objective Function}}: Given a document $d_b$, a target model $f$ and hyper-parameters $\gamma, \delta$, $\epsilon$, $k$, our goal is to find a perturbed document $d_p$ that satisfies the following:
\begin{equation}\label{problem}
    \begin{aligned}
    &\mathbf{\textbf{\textit{Sim}}_e}(e_{d_b},e_{d_p}) > \gamma, \\ \\
        \mathrm{s.t.} \;\;&f(d_b) = f(d_p), \\
        &\mathbf{\textbf{\textit{Sim}}_s}(d_b,d_p) \geq \delta, \\
        &i \leq \epsilon * |f|, \\
        &e_{d_p} \cap c \: \neq \: \varnothing\;\;\; \forall \: c \in e_{d_b}[:k]
    \end{aligned}
\end{equation}
}}
\end{center}

\subsection{Similarity  Measures}\label{sec:sim_measures}
Since the similarity measure is the engine that drives the comparison process we need to choose measures carefully to provide useful results.
We assume the explanations generated are ranked lists, ordered by importance to the surrogate model (as is standard for \textsc{Lime} and common with other XAI methods). It is natural to then choose measures for similarity or distance designed specifically for ranked lists. We use the candidates (and the success thresholds of 30, 40, 50, and 60\%) provided in the concurrent work on similarity measures in XAI \autocite{burger2024effectsimilaritymeasuresaccurate}. Briefly, these measures are: The Jaccard Index (The ratio of the intersection over the union), Rank-biased Overlap (Summation of intersections of increasing size with weighting applied via a parameter (denoted in subscript) controlled convergent series, where values approach one assign more weight the topmost features), Kendall's Tau Rank Distance (Number of dissonant pairs of features as ordered by index), and Spearman's footrule (Effectively the $l_1$ distance between explanations). All non-RBO measures also have a weighted alternative based on the values calculated by the explanatory model. See \cite{burger2024effectsimilaritymeasuresaccurate} for more detail on the definitions and features of these measures.

\subsection{Search Procedure}
The general search process is focused on the comparison of explanations, and not on how the perturbations are generated. Again, we appeal to previous work which has used a greedy search where the indices of the original document $d_b$ are ordered by importance (or lack thereof) to the base model $f$. This importance is determined by calculating the difference in output probability when the word at index $i$ is removed. The sorted indices are then iterated through until perturbation limit or exhaustion replacing the word at index $i$ with its $j$ nearest neighbors in some embedding space and choosing the replacement with the largest similarity decrease. 

As minimal perturbations were not the focus of these search processes, we additionally test a genetic algorithm approach for the selection of location and choice of word replacement. Genetic algorithms efficiently search extremely large search spaces, making them a good candidate for efficiently searching for minimal perturbations. Prior work in non-XAI adversarial examples for natural language have used genetic algorithms to good effect in producing parsimonious replacements \parencite{alzantot2018generating}. As our goal is to explore the lower bounds of required perturbations to achieve a desired similarity difference, it is natural to include this alternative search method.

\newcolumntype{Y}{>{\centering\arraybackslash}X}
\begin{table*}[ht]
\caption{Attack Success Rates}
\label{tab:attack_success}
\footnotesize
\begin{tabularx}{\textwidth}{c *{20}{Y}}
\toprule
\multicolumn{1}{c}{\textbf{}}
& \multicolumn{1}{c}{\textbf{$\tau$}}
& \multicolumn{2}{c}{\textbf{RBO}$_{0.5}$}
& \multicolumn{2}{c}{\textbf{RBO}$_{0.7}$}
& \multicolumn{2}{c}{\textbf{RBO}$_{0.9}$}
& \multicolumn{2}{c}{\textbf{Jaccard}}
& \multicolumn{2}{c}{\textbf{Jaccard}$_w$}
& \multicolumn{2}{c}{\textbf{Kendall}}
& \multicolumn{2}{c}{\textbf{Kendall}$_w$}
& \multicolumn{2}{c}{\textbf{Spearman}}
& \multicolumn{2}{c}{\textbf{Spearman}$_w$}\\
\cmidrule(lr){2-20}
& & \textbf{GA} & \textbf{GS} & \textbf{GA} & \textbf{GS} & \textbf{GA} & \textbf{GS} & \textbf{GA} & \textbf{GS} & \textbf{GA} & \textbf{GS} & \textbf{GA} & \textbf{GS} & \textbf{GA} & \textbf{GS} & \textbf{GA} & \textbf{GS} & \textbf{GA} & \textbf{GS} \\ \addlinespace[3pt]
& 30\%  & 0.24 &  0.15 & 0 & 0 & 0 & 0 & 0  & 0   & 0 & 0 &  1 &  0.95 & 0.53 & 0.65 & 0.12  & 0.20  & 0.24  & 0\\
\rotatebox[origin=c]{90}{\textbf{GB}}& 40\%   &  0.35 & 0.25  & 0.24 & 0.05 & 0.06  & 0 & 0.12 & 0.20   & 0 &  0 &  1 & 1 & 0.65 & 0.75  & 0.29  & 0.35  & 0.35 & 0.15\\
& 50\%  & 0.47 &  0.45 &  0.35 & 0.30 & 0.12 & 0.05 & 0.82 & 0.85 & 0.06 & 0 & 1 &  1 &  0.82 & 0.85 & 0.82 & 0.80  & 0.48  & 0.35 \\
& 60\%  & 0.47&  0.45 &  0.53 & 0.45 & 0.41 & 0.35 &  1 & 1 & 0.29    & 0 & 1 &  1 &  0.88 & 0.90 & 1 & 1  & 0.88  & 0.65 \\
  \midrule

& 30\%  & 0.45 &  0.09 & 0.15 & 0 & 0 & 0 & 0.05  & 0.09   & 0 & 0 &  1 &  1 & 0.65 & 0.30 & 0.15  & 0.18  & 0.35  & 0.04\\
\rotatebox[origin=c]{90}{\textbf{S2D}}& 40\%   &  0.45 & 0.13  & 0.30 & 0.04 & 0.10  & 0 & 0.50 & 0.65   & 0.10 &  0 &  1 & 1 & 0.70 & 0.43  & 0.50  & 0.52  & 0.50 & 0.30\\
& 50\%  & 0.50 &  0.22 &  0.40 & 0.13 & 0.35 & 0.13 & 0.95 & 1    & 0.30 & 0.04 &  1 &  1 & 0.90 & 0.78 & 0.90  & 0.87  & 0.80  & 0.61\\
& 60\%  & 0.50&  0.22 &  0.50 & 0.26 & 0.65 & 0.39 & 1 & 1    & 0.40 & 0.18 &  1 &  1 & 1 & 0.95 & 1  & 1  & 0.85  & 0.87\\
\bottomrule
\end{tabularx}
\end{table*}


\begin{table*}[ht]
\caption{Mean Similarities}
\label{tab:mean_similarities}
\footnotesize
\begin{tabularx}{\textwidth}{c *{20}{Y}}
\toprule
\multicolumn{1}{c}{\textbf{}}
& \multicolumn{1}{c}{\textbf{$\tau$}}
& \multicolumn{2}{c}{\textbf{RBO}$_{0.5}$}
& \multicolumn{2}{c}{\textbf{RBO}$_{0.7}$}
& \multicolumn{2}{c}{\textbf{RBO}$_{0.9}$}
& \multicolumn{2}{c}{\textbf{Jaccard}}
& \multicolumn{2}{c}{\textbf{Jaccard}$_w$}
& \multicolumn{2}{c}{\textbf{Kendall}}
& \multicolumn{2}{c}{\textbf{Kendall}$_w$}
& \multicolumn{2}{c}{\textbf{Spearman}}
& \multicolumn{2}{c}{\textbf{Spearman}$_w$}\\
\cmidrule(lr){2-20}
& & \textbf{GA} & \textbf{GS} & \textbf{GA} & \textbf{GS} & \textbf{GA} & \textbf{GS} & \textbf{GA} & \textbf{GS} & \textbf{GA} & \textbf{GS} & \textbf{GA} & \textbf{GS} & \textbf{GA} & \textbf{GS} & \textbf{GA} & \textbf{GS} & \textbf{GA} & \textbf{GS} \\ \addlinespace[3pt]
& 30\%  & 0.26 &  0.30 & - & - & - & - & -  & -   & - & - &  0.18 &  0.18 & 0.11 & 0.11 & 0.23  & 0.26  & 0.19  & -\\
\rotatebox[origin=c]{90}{\textbf{GB}}& 40\%   &  0.33 & 0.34  & 0.36 & 0.39 & 0.38  & - & 0.36 & 0.38   & - &  - &  0.24 & 0.23 & 0.20 & 0.18  & 0.34  & 0.34  & 0.25 & 0.38\\
& 50\%  & 0.41 &  0.39 &  0.45 & 0.48 & 0.44 & 0.44 & 0.48 & 0.45    & 0.44 & 0 &  0.27 &  0.31 & 0.29 & 0.24 & 0.45  & 0.48  & 0.41  & 0.47\\
& 60\%  & 0.41&  0.29 &  0.55 & 0.53 & 0.55 & 0.57 & 0.54 & 0.55    & 0.53 & 0 &  0.31 &  0.33 & 0.35 & 0.27 & 0.53  & 0.56  & 0.54  & 0.54\\
  \midrule

& 30\%  & 0.24 &  0.29 & 0.23 & - & - & - & 0.25  & 0.29   & - & - &  0.18 &  0.23 & 0.15 & 0.22 & 0.27  & 0.25  & 0.22  & 0.22\\
\rotatebox[origin=c]{90}{\textbf{S2D}}& 40\%   &  0.25 & 0.34  & 0.35 & 0.39 & 0.37  & - & 0.37 & 0.37   & 0.37 &  0 &  0.21 & 0.28 & 0.23 & 0.32  & 0.37  & 0.37  & 0.29 & 0.38\\
& 50\%  & 0.34 &  0.43 &  0.41 & 0.49 & 0.48 & 0.49 & 0.46 & 0.47    & 0.46 & 0.47 &    0.23 & 0.36 & 0.33 & 0.42  & 0.46  & 0.47  & 0.38 & 0.45\\
& 60\%  & 0.34&  0.43 &  0.48 & 0.55 & 0.56 & 0.59 & 0.56 & 0.56    & 0.54 & 0.55 &  0.23 &  0.38 & 0.41 & 0.50 & 0.53  & 0.56  & 0.53  & 0.57\\
\bottomrule
\end{tabularx}
\end{table*}

\section{Genetic Algorithm}

Genetic algorithms utilize the concept of digital evolution to 'evolve' solutions to are composed of a population of candidate solutions (chromosomes). The population is then put through parent selection, crossover, and mutation for a number of generations until a stopping condition has been met. For each generation, a fitness function is used to evaluate the quality of each candidate solution. 
Our genetic algorithm follows similarly to Alzantot et al. (\citeyear{alzantot2018generating})

We initially define a fitness function, allowing the algorithm to determine the highest quality chromosomes through the evolution process. For our purposes, the fitness function is the Objective Function (\ref{problem}).

We then define the mutation function. All aspects of the mutation function are subject to the constraints in Section \ref{sec:perturbation_strategy}.

\begin{enumerate}
    \item Collect all indices valid for perturbation
    \item Select an arbitrary index and generate replacements for the word at this location
    \item Test replacements against the prior best candidate
    \item If superior, update new best candidate
    \item Repeat until exhaustion or success
\end{enumerate}

With the mutation algorithm, we generate an initial population, mutating the base document $n$ times. We then order the population by fitness, truncating half the lowest quality chromosomes and ensuring the highest quality mutations are selected for continued modification. Once a set of parents are selected, crossover is performed by selecting two arbitrary members of the population, selecting an arbitrary index $i$ and swapping the remainder of the text past index $i$ between these members. If no valid child was generated from the previous two steps, we choose an arbitrary parent as the child. Once the number of valid children matches the size of the parent population, we mutate all children. The child population is then used as the parent population for the next generation of the algorithm.

Our genetic algorithm uses a population of size 10 and a maximum number of generations of 10. These limits are restrictive compared to most other genetic algorithms, but computational limitations prevented any appreciable increase. The genetic algorithm as it exists here is already subject to $\sim 250\%$ increase in time taken per example over the greedy search.

\section{Experiment}
We test the alternate search process in finding more efficient perturbations by performing a comparison against the greedy search method in Burger et al. (\citeyear{xaifooler}). We use two of the datasets in the previous paper, the short length Twitter Gender Bias (\textbf{GB}) with an average of 11 words and the intermediate length Symptoms to Diagnosis (\textbf{S2D} with an average of 29 words. We were unable to test the collection of similarity measures and thresholds on larger datasets ($\geq 100$ words) due to the computational expenditure required. From these datasets we select twenty examples to perturb across the range of thresholds and similarity measures described in Section \ref{sec:sim_measures} resulting in a total of 720 examples per dataset or 1440 total examples. From these attacks we compare the proportion of successful attacks (Table \ref{tab:attack_success}), end similarity for successful attacks (Table \ref{tab:mean_similarities}), proportion of words perturbed for successful attacks (Table \ref{tab:average_perturbation_rate}), and minimum required number of perturbations to successfully attack any example within the dataset (Table \ref{tab:minimum_perturbations}).

\begin{table*}[h]
\caption{Average Perturbation Rate for Successful Attacks. 
}
\footnotesize
\begin{tabularx}{\textwidth}{c *{20}{Y}}
\toprule
\multicolumn{1}{c}{\textbf{}}
& \multicolumn{1}{c}{\textbf{$\tau$}}
& \multicolumn{2}{c}{\textbf{RBO}$_{0.5}$}
& \multicolumn{2}{c}{\textbf{RBO}$_{0.7}$}
& \multicolumn{2}{c}{\textbf{RBO}$_{0.9}$}
& \multicolumn{2}{c}{\textbf{Jaccard}}
& \multicolumn{2}{c}{\textbf{Jaccard}$_w$}
& \multicolumn{2}{c}{\textbf{Kendall}}
& \multicolumn{2}{c}{\textbf{Kendall}$_w$}
& \multicolumn{2}{c}{\textbf{Spearman}}
& \multicolumn{2}{c}{\textbf{Spearman}$_w$}\\
\cmidrule(lr){2-20}
& & \textbf{GA} & \textbf{GS} & \textbf{GA} & \textbf{GS} & \textbf{GA} & \textbf{GS} & \textbf{GA} & \textbf{GS} & \textbf{GA} & \textbf{GS} & \textbf{GA} & \textbf{GS} & \textbf{GA} & \textbf{GS} & \textbf{GA} & \textbf{GS} & \textbf{GA} & \textbf{GS} \\ \addlinespace[3pt]
& 30\%  & 0.18 &  0.12 & - & - & - & - & -  & -   & - & - &  0.12 &  0.11 & 0.12 & 0.09 & 0.19  & 0.20  & 0.21  & -\\
\rotatebox[origin=c]{90}{\textbf{GB}}& 40\%   &  0.17 & 0.14  & 0.21 & 0.19 & 0.26  & - & 0.26 & 0.22   & - &  - &  0.10 & 0.10 & 0.11 & 0.09  & 0.18  & 0.20  & 0.20 & 0.22\\
& 50\%  & 0.07 &  0.07 &  0.18 & 0.16 & 0.26 & 0.25 & 0.22 & 0.22    & 0.27 & - &  0.08 &  0.08 & 0.09 & 0.09 & 0.18  & 0.17  & 0.16  & 0.15\\
& 60\%  & 0.07&  0.07 &  0.11 & 0.09 & 0.23 & 0.23 & 0.19 & 0.17    & 0.23 & - &  0.08 &  0.08 & 0.09 & 0.09 & 0.14  & 0.14  & 0.15  & 0.15\\
  \midrule

& 30\%  & 0.11 &  0.11 & 0.18 & - & - & - & 0.24  & 0.23   & - & - &  0.05 &  0.06 & 0.09 & 0.07 & 0.16  & 0.17  & 0.14  & 0.15\\
\rotatebox[origin=c]{90}{\textbf{S2D}}& 40\%   &  0.11 & 0.12  & 0.13 & 0.14 & 0.20  & - & 0.20 & 0.20   & 0.16 &  - &  0.04 & 0.05 & 0.08 & 0.06  & 0.15  & 0.17  & 0.14 & 0.13\\
& 50\%  & 0.08 &  0.04 &  0.10 & 0.10 & 0.19 & 0.20 & 0.18 & 0.17    & 0.17 & 0.21 &  0.04 &  0.04 & 0.07 & 0.07 & 0.14  & 0.14  & 0.13  & 0.14\\
& 60\%  & 0.08 &  0.04 &  0.09 & 0.07 & 0.17 & 0.17 & 0.14 & 0.12    & 0.15 & 0.20 &  0.04 &  0.04 & 0.06 & 0.06 & 0.12  & 0.10  & 0.11  & 0.11\\
\bottomrule
\end{tabularx}

\label{tab:average_perturbation_rate}
\end{table*}

\begin{table*}[h]
\caption{Minimum Perturbation(s) required for a successful attack }
\footnotesize
\begin{tabularx}{\textwidth}{c *{20}{Y}}
\toprule
\multicolumn{1}{c}{\textbf{}}
& \multicolumn{1}{c}{\textbf{$\tau$}}
& \multicolumn{2}{c}{\textbf{RBO}$_{0.5}$}
& \multicolumn{2}{c}{\textbf{RBO}$_{0.7}$}
& \multicolumn{2}{c}{\textbf{RBO}$_{0.9}$}
& \multicolumn{2}{c}{\textbf{Jaccard}}
& \multicolumn{2}{c}{\textbf{Jaccard}$_w$}
& \multicolumn{2}{c}{\textbf{Kendall}}
& \multicolumn{2}{c}{\textbf{Kendall}$_w$}
& \multicolumn{2}{c}{\textbf{Spearman}}
& \multicolumn{2}{c}{\textbf{Spearman}$_w$}\\
\cmidrule(lr){2-20}
& & \textbf{GA} & \textbf{GS} & \textbf{GA} & \textbf{GS} & \textbf{GA} & \textbf{GS} & \textbf{GA} & \textbf{GS} & \textbf{GA} & \textbf{GS} & \textbf{GA} & \textbf{GS} & \textbf{GA} & \textbf{GS} & \textbf{GA} & \textbf{GS} & \textbf{GA} & \textbf{GS} \\ \addlinespace[3pt]
& 30\%  & 1 &  1 & - & - & - & - & -  & -   & - & - &  1 &  1 & 1 & 1 & 3  & 2  & 2  & -\\
\rotatebox[origin=c]{90}{\textbf{GB}}& 40\%   &  1 & 1  & 2 & 3 & 4  & - & 4 & 3   & - &  - &  1 & 1 & 1 & 1  & 2  & 2  & 2 & 3\\
& 50\%  & 1 &  1 &  1 & 1 & 4 & 4 & 2 & 2    & 4 & - &  1 &  1 & 1 & 1 & 2  & 1  & 1  & 1\\
& 60\%  & 1 &  1 &  1 & 1 & 3 & 2 & 2 & 2    & 2 & - &  1 &  1 & 1 & 1 & 1  & 1  & 1  & 1\\
  \midrule

& 30\%  & 1 &  2 & 4 & - & - & - & 5  & 6   & - & - &  1 &  1 & 1 & 1 & 4  & 4  & 3  & 7\\
\rotatebox[origin=c]{90}{\textbf{S2D}}& 40\%   &  1 & 1  & 3 & 4 & 6  & - & 3 & 4   & 3 &  - &  1 & 1 & 1 & 1  & 2  & 2  & 2 & 2\\
& 50\%  & 1 &  1 &  1 & 1 & 5 & 5 & 3 & 3    & 3 & 6 &  1 &  1 & 1 & 1 & 1  & 2  & 1  & 2\\
& 60\%  & 1&  1 &  1 & 1 & 3 & 4 & 2 & 2 & 3 & 6 &  1 &  1 & 1 & 1 & 1  & 1  & 1  & 1\\
\bottomrule
\end{tabularx}

\label{tab:minimum_perturbations}
\end{table*}

\begin{figure}[htb]
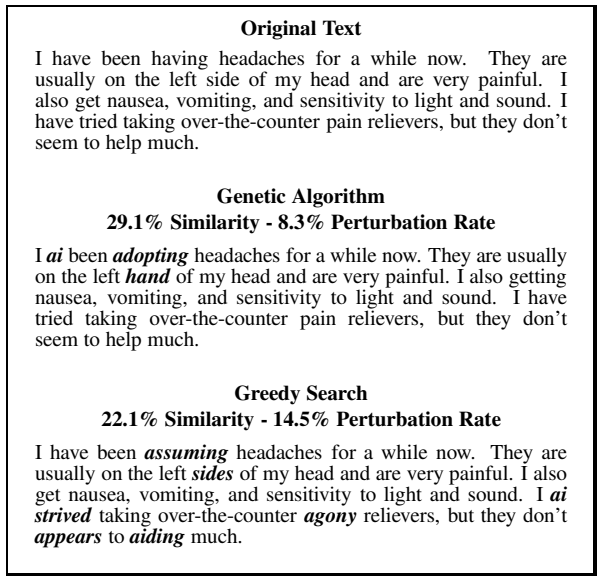

    \footnotesize
    \begin{mdframed}
    
     \begin{center}
     \textbf{Original Text}   
    \end{center}
        \vspace{3pt}

    I have been having headaches for a while now. They are usually on the left side of my head and are very painful. I also get nausea, vomiting, and sensitivity to light and sound. I have tried taking over-the-counter pain relievers, but they don't seem to help much.
    \newline

        \begin{center}
     \textbf{Genetic Algorithm \\ \vspace{2pt} 29.1\% Similarity - 8.3\% Perturbation Rate }
    \end{center}

     I \textbf{\textit{ai}} been \textbf{\textit{adopting}} headaches for a while now. They are usually on the left \textbf{\textit{hand}} of my head and are very painful. I also getting nausea, vomiting, and sensitivity to light and sound. I have tried taking over-the-counter pain relievers, but they don't seem to help much.
    \newline

        \begin{center}
     \textbf{Greedy Search \\ \vspace{2pt} 22.1\% Similarity - 14.5\% Perturbation Rate}
    \end{center}

    I have been \textbf{\textit{assuming}} headaches for a while now. They are usually on the left \textbf{\textit{sides}} of my head and are very painful. I also get nausea, vomiting, and sensitivity to light and sound. I \textbf{\textit{ai}} \textbf{\textit{strived}} taking over-the-counter \textbf{\textit{agony}} relievers, but they don't \textbf{\textit{appears}} to \textbf{\textit{aiding}} much.
    
\vspace{5pt}

\end{mdframed} 
\caption{Successful attacks at the 30\% threshold under similarity measure Spearman$_w$. 
}
\label{fig:perturbation_reduction}
  
\end{figure}

\section{Results}

The genetic algorithm fulfills our goal of providing a way to better explore the large possible search space. For the \textbf{GB} dataset we find a substantial increases in successful attacks under Spearman$_w$, especially at the 30\% threshold ($0 \to 24\%$). Similarly, we also see successful attacks where there were none under Jaccard$_w$ ($0 \to 6\%\;  \tau = 50\%, \; 0 \to 29\%\;  \tau = 60\% $) under the greedy search (\textbf{GS}) (Table \ref{tab:attack_success}). The average perturbation rates, minimum perturbations and ending similarity remain close to the greedy search showing the genetic algorithm still offers comparable performance across other measures.

For the \textbf{S2D} dataset we see results comparable to the \textbf{GB} dataset with the Jaccard$_w$ and Spearman$_w$ measures. For these we see the genetic algorithm has the capacity to produce explanations with substantially fewer perturbations that still satisfy a particular threshold (Table \ref{tab:minimum_perturbations}). Attacks on RBO in general were also more successful over the greedy search but with a less pronounced difference in minimal perturbations.

The importance of emphasizing minimal perturbations can be seen in Figure \ref{fig:perturbation_reduction}. The word replacement algorithm (TextFooler (\cite{Textfooler})) can struggle with appropriate replacements despite all the constraints applied, and we see here that no individual perturbation here is particularly good. The best way to mitigate this is then to reduce the overall number of perturbations. Ideally the perturbations would be both inconspicuous and minimal, but as adversarial XAI research is still in its early stages we must use existing tools until superior versions can be developed.

\begin{figure}[thb]
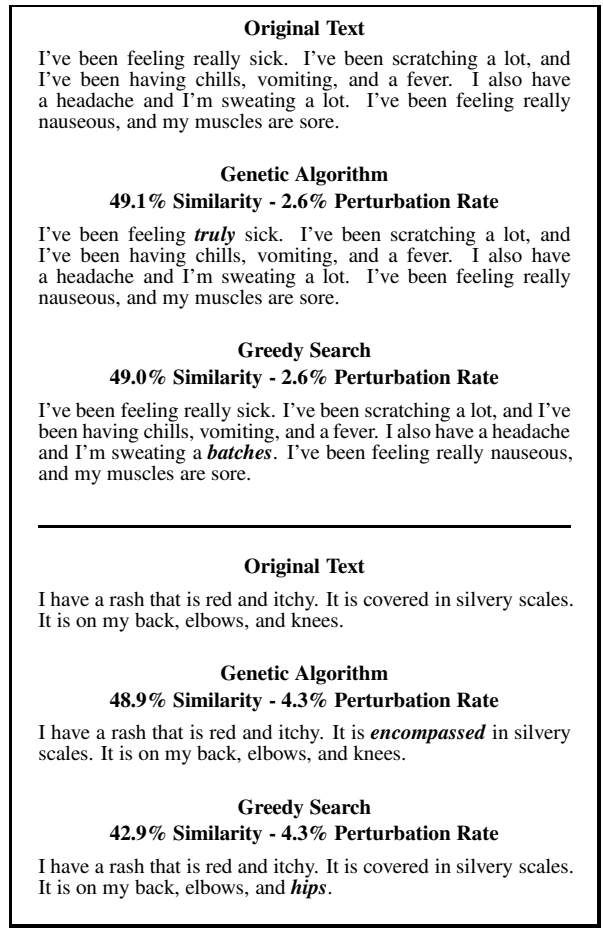

    \footnotesize
    \begin{mdframed}
     \begin{center}
     \textbf{Original Text}   
    \end{center}
        \vspace{3pt}

      I've been feeling really sick. I've been scratching a lot, and I've been having chills, vomiting, and a fever. I also have a headache and I'm sweating a lot. I've been feeling really nauseous, and my muscles are sore.
    \newline

        \begin{center}
     \textbf{Genetic Algorithm \\ \vspace{2pt} 49.1\% Similarity - 2.6\% Perturbation Rate }
    \end{center}

       I've been feeling \textbf{\textit{truly}} sick. I've been scratching a lot, and I've been having chills, vomiting, and a fever. I also have a headache and I'm sweating a lot. I've been feeling really nauseous, and my muscles are sore.
    \newline

        \begin{center}
     \textbf{Greedy Search \\ \vspace{2pt} 49.0\% Similarity - 2.6\% Perturbation Rate}
    \end{center}

      I've been feeling really sick. I've been scratching a lot, and I've been having chills, vomiting, and a fever. I also have a headache and I'm sweating a \textbf{\textit{batches}}. I've been feeling really nauseous, and my muscles are sore.
    
\vspace{10pt}

\hrulefill

\vspace{5pt}

     \begin{center}
     \textbf{Original Text}   
    \end{center}
    
      I have a rash that is red and itchy. It is covered in silvery scales. It is on my back, elbows, and knees.
    \newline

        \begin{center}
     \textbf{Genetic Algorithm \\ \vspace{2pt} 48.9\% Similarity - 4.3\% Perturbation Rate }
    \end{center}

       I have a rash that is red and itchy. It is \textbf{\textit{encompassed}} in silvery scales. It is on my back, elbows, and knees.
    \newline

        \begin{center}
     \textbf{Greedy Search \\ \vspace{2pt} 42.9\% Similarity - 4.3\% Perturbation Rate}
    \end{center}

      I have a rash that is red and itchy. It is covered in silvery scales. It is on my back, elbows, and \textbf{\textit{hips}}.
    
\vspace{5pt}
\end{mdframed} 

\caption{Successful attacks at the 50\% threshold under similarity measure RBO$_{0.5}$. The genetic algorithm can produce perturbations with worse (higher) similarity but with greater textual quality.}
\label{fig:superior_single_perturbations}
\end{figure}

The genetic algorithm can produce improvements apart from minimal perturbations. In Figure \ref{fig:superior_single_perturbations} we see two examples of single perturbation successes. Both the greedy search and genetic algorithm produce a successful attack, but the perturbed word differs. The first example has the greedy search change the word \textbf{lot} to \textbf{batches}, which while naively reasonable in terms of the similarity between the words, it lacks appropriate context and so degrades the quality of the perturbed document. The genetic algorithm however replaces \textbf{really} with \textbf{truly}, which maintains the meaning and structure of the original text completely. 


The second example has the greedy search replace \textbf{knees} with \textbf{hips}. While \textbf{knees} and \textbf{hips} are close neighbors in the embedding space used to locate the perturbations, this significantly changes the meaning of the text despite satisfying all of the constraints and maintaining the overall quality of the text. The genetic algorithm replaces \textbf{covered} with \textbf{encompassed}, which provides a fairly inconspicuous replacement while reaching the set threshold.
\newpage

However, the genetic algorithm is not strictly superior to the greedy search. In Figure \ref{fig:inferior_perturbation} the greedy search generates an ideal perturbation of \textbf{good} to \textbf{nice} resulting in perfectly maintained meaning while the genetic algorithm's replacement is stilted and cumbersome (\textbf{hit} to \textbf{strike}).






    


\begin{figure}[t]
    \footnotesize
    \begin{mdframed}
     \begin{center}
     \textbf{Original Text}   
    \end{center}
    \vspace{3pt}
      you hit the two top words on your first try . . . good job ! yar and matey
    \newline

        \begin{center}
     \textbf{Genetic Algorithm \\ \vspace{2pt} 43.1\% Similarity - 6.7\% Perturbation Rate }
    \end{center}

       you \textbf{\textit{strike}} the two top words on your first try . . . good job ! yar and matey
    \newline

        \begin{center}
     \textbf{Greedy Search \\ \vspace{2pt} 41.5\% Similarity - 6.7\% Perturbation Rate}
    \end{center}

      you hit the two top words on your first try . . . \textbf{\textit{nice}} job ! yar and matey
    
\vspace{5pt}
\end{mdframed} 

\caption{The greedy search can produce a superior perturbation with respect to both similarity and quality.}
\label{fig:inferior_perturbation}
\end{figure}

\section{Limitations and Conclusion}
Overall, the alternate search process used has demonstrated effectiveness in both reducing the number of perturbations, while having the capacity to meet or exceed the original greedy search in terms of perturbed document quality and attack success rates. Given the relative simplicity of the genetic algorithm and the large possible variety of parent selection methods, mutation strategies and chromosome representations available, further increases in efficacy seem likely. Unfortunately, the extensive computational requirements make testing these many combinations a difficult task. Any explanatory method used will likely require a purpose built adaptation to allow efficient computation on modern hardware, especially as existing methodologies are lacking in optimal GPU support. But once such methods are developed the conclusions about the vulnerability of XAI methods to adversarial attack will be that much stronger.

\section{Ethical Considerations}
We do not anticipate any adverse ethical considerations from this work. As the focus of this work is on establishing a more effective understanding of the limitations of XAI methods, we expect it to promote accountability and correctness in future undertaking in adversarial XAI.

\section{Acknowledgements}
Christopher Burger was partially funded with a fellowship from the University of Mississippi's Institute for Data Science.



\printbibliography

\end{document}